\documentclass[11pt]{article}
\usepackage{graphicx}
\usepackage{amssymb}
\usepackage{amscd}
\usepackage{mathrsfs}
\usepackage{longtable,lscape}
\usepackage{amsthm}
\usepackage{amsfonts}
\usepackage{amsmath}
\usepackage{bbm}
\usepackage{float}
\usepackage{url}

\newcommand{\captionfonts}{\footnotesize}
\makeatletter  
\long\def\@makecaption#1#2{%
  \vskip\abovecaptionskip
  \sbox\@tempboxa{{\captionfonts #1: #2}}%
  \ifdim \wd\@tempboxa >\hsize
    {\captionfonts #1: #2\par}
  \else
    \hbox to\hsize{\hfil\box\@tempboxa\hfil}%
  \fi
  \vskip\belowcaptionskip}
\makeatother 
\begin{document}
\title{Quantum Structure in Cognition and the Foundations \\ of Human Reasoning}
\author{Diederik Aerts$^1$, Sandro Sozzo$^{2,1}$ and Tomas Veloz$^{3,1}$ \vspace{0.5 cm} \\ 
        \normalsize\itshape
        $^1$ Center Leo Apostel for Interdisciplinary Studies \\
        \normalsize\itshape
        and, Department of Mathematics, Brussels Free University \\ 
        \normalsize\itshape
         Krijgskundestraat 33, 1160 Brussels, Belgium \\
        \normalsize
        E-Mails: \url{diraerts@vub.ac.be, ssozzo@vub.ac.be}
          \vspace{0.5 cm} \\ 
        \normalsize\itshape
        $^2$ School of Management and IQSCS, Universit
        y of Leicester \\ 
        \normalsize\itshape
         University Road, LE1 7RH Leicester, United Kingdom \\
        \normalsize
        E-Mail: \url{ss831@le.ac.uk}
          \vspace{0.5 cm} \\ 
        \normalsize\itshape
        $^3$ Department of Mathematics, University of British Columbia, Okanagan Campus \\
        \normalsize\itshape
        3333 University Way, Kelowna, BC Canada V1V 1V7
        \\
        \normalsize\itshape
        and, Instituto de Filosof\'ia y Ciencias de la Complejidad (IFICC)  \\
        \normalsize\itshape
        Los Alerces 3024, \~Nu\~noa, Santiago, Chile  \\
        \normalsize
        E-Mail: \url{tomas.veloz@ubc.ca} \\
              }
\date{}
\maketitle
\begin{abstract}
\noindent
Traditional
cognitive science rests on a foundation of classical logic and probability theory. This foundation has been seriously challenged by several findings in experimental psychology on human decision making. Meanwhile, the formalism of quantum theory has provided an efficient resource for modeling these classically problematical situations. In this paper, we start from our successful quantum-theoretic approach to the modeling of concept combinations to formulate a unifying explanatory hypothesis. In it, human reasoning is the superposition of two processes -- a conceptual reasoning, whose nature is emergence
 of new conceptuality, and a logical reasoning, 
founded on an algebraic calculus of the logical type. In most cognitive processes however, the former reasoning prevails over the latter. In this perspective, the observed
deviations from classical logical reasoning should not be interpreted as biases but, rather, as natural expressions of emergence in its deepest form.
\end{abstract}

\section{Introduction\label{intro}}
Concepts are envisaged as the structural units of human thought.  Understanding the nature of these units can lead to a first principles basis for a foundational theory of cognition, especially for the study of decision making, problem solving, and communication~\cite{r1973,g2000}. Concepts can be used to represent very different forms of meaning, ranging from concrete objects (e.g., ``this cup of coffee in front of me'') to abstract forms of experience (e.g., ``absolute free will'').
Although concepts have been conceived to represent classes of similar items, it is generally accepted that 
they do not have a fixed representational structure~\cite{f1998}. 
Concretely, this 
view 
is based on three conceptual phenomena. The first is `vagueness', i.e. concepts have no sharp boundaries~\cite{h2007}. The second is `contextuality', i.e. concepts acquire meaning depending on the situation they are elicited~\cite{r1976}. These two phenomena have been accounted for, at least to some extent, by fuzzy set theory and classical probability theory \cite{z1989}. The third phenomenon corresponds to `concept combination', i.e. how 
the meaning of combined concepts 
relates to the meaning of their constituents. This phenomenon has been vastly investigated by psychologists and cognitive scientists~\cite{r1995,kp1995,h1997}. Experiments in concept combination 
traditionally request participants to estimate 
typicality or membership of certain items of concepts and their combinations. The goal is to obtain a model where 
typicality or membership of the combined concept 
is related to typicality or memberships of the constituent concepts. It has been shown that experimental estimates of 
typicalities \cite{os1981} and of memberships 
\cite{a2009a} cannot be modeled respectively within fuzzy logics and classical probability. 
In particular, given two concepts $A$ and $B$, and an item $x$, the membership of $x$ with respect to a combined concept `$A$ and $B$' (`$A$ or $B$') is very often larger (smaller) than the membership of $x$ with respect to both $A$ and $B$. This contradicts fuzzy and classical probabilistic 
rules~\cite{os1981,h1988a,h1988b}.

In this paper we support the view that a novel theoretical framework to cope with the problems of concept combination is needed. In particular, we have developed a modeling approach for concept combination that uses the mathematical formalism of quantum theory \cite{a2009a,ag2005a,ag2005b,as2011,ags2013,s2014b,asv2014b}. This approach enables faithful representation of a large amount of data collected on conjunctions and disjunctions of two concepts (Sect. \ref{conjdisj}) and, more recently, conjunctions and negations (Sect. \ref{CC-Neg}). Starting from the success of our quantum-theoretic approach, we formulate in Sect. \ref{nature} an explanatory hypothesis, according to which two types of reasoning simultaneously occur in a human decision making (conceptual membership estimation, human probability judgement, gamble preference under uncertainty, etc.) and, more generally, in a cognitive process. The first type is `emergent reasoning', the second is `logical reasoning'. This hypothesis is sustained and justified in Sects. \ref{conjdisj} and \ref{CC-Neg} by considering relevant highlights in concept theory. Finally, we investigate in Sect. \ref{aristotle} the implications of our explanatory hypothesis on the role played by the observed deviations from classicality in human decision making, and elaborate some important insights into the origin of logic and the nature of human reasoning.   

\section{An explanatory hypothesis on the nature of human reasoning\label{nature}}

We elaborate in this section our explanation for the 
appearance of genuine quantum structures in cognitive processes, which has matured during our research on conceptual combination and human decision making. This explanation reveals very stable patterns of human 
reasoning, enlightening
at the same time
some fundamental traits of its deepest nature.

That the fundamental nature of human reasoning should include probabilistic aspects was already known to cognitive psychologists at the beginning of the seventies, when Rosch's experiments confirmed that the combination of natural concepts in human thought exhibits `graded typicality', or `vagueness'. However, one  believed that the observed probabilities could be cast into the Kolmogorovian
  structures of classical probability theory, thus revealing an underlying classical logical behaviour, possibly extended to include fuzzy set logic and its basic connectives. This was the `admirable illusion' of cognitive psychology in the last century.

This `human rational behaviour myth' was put at stake by different discoveries in concept theory. Osherson and Smith observed that human subjects estimate some items to be more typical examples of the concept conjunction `$A$ and $B$' than of $A$ and $B$, separately \cite{os1981}. Hampton observed that, for several items, the `membership weight', that is, the degree of memberhip, of the item with respect to the conjunction `$A$ and $B$' is generally higher than its membership weight with respect to $A$ or $B$, while the membership weight with respect to the disjunction `$A$ or $B$' is generally lower than the membership weight with respect to $A$ or $B$ \cite{h1988a,h1988b}. In the same period, the so-called `Tversky and Khaneman program' revealed the `disjunction effect' and the `conjunction fallacy' in human decision making \cite{tk1983,ts1992}, which has a counterpart  in the `Allais', `Ellsberg' and `Machina paradoxes' of behavioural economics \cite{e1961,m2009}. The general attitude was then to consider these deviations from classicality as `fallacies',  as `effects', as `paradoxes', or as `contradictions'. In other words, traditional approaches to cognition interpreted such deviations as `biases of classical logical reasoning'. 

A different approach to cognitive psychology, initiated two decades ago, has meanwhile matured into a new domain of research, called `quantum cognition'. Its main feature is the use of the mathematical formalism of quantum theory as modeling tool for these cognitive situations where traditional classically based approaches fail. Important results have been obtained in the quantum modeling of the above decision making situations. Specifically, we have investigated the dynamics of concepts and how they combine as our contribution to quantum cognition, and were able to represent a huge amount of data collected in different experiments on conceptual combinations in a 
quantum-mechanical framework.  
Our quantum-theoretic approach on concepts and their combinations leads us to formulate a specific hypothesis on the mechanisms that underlie human 
reasoning, not only when it combines concepts to form sentences and texts, or in a decision process, but, more generally, in any cognitive process.

According to our explanatory hypothesis, human reasoning is a specifically structured superposition of two processes, a `logical reasoning' and a `emergent reasoning'. The former 
`logical reasoning' combines cognitive entities, such as concepts,
combinations of concepts, or propositions,  by applying the rules of logic, though 
generally in a probabilistic 
way. The latter 
`emergent reasoning' enables formation of combined cognitive entities as 
newly emerging 
entities, in the case of concepts, new concepts, in the case of propositions, new propositions, 
carrying new meaning, linked to the meaning of the constituent cognitive entities, but with a linkage not defined by the algebra of logic.
The two mechanisms act simultaneously and in superposition in human thought 
during a reasoning process, the first one is guided by an algebra of `logic', the second one 
follows a mechanism of `emergence'. In this perspective, human reasoning can be mathematically formalized in a two-sector Fock space, where 
the states of conceptual entities are represented 
by unit vectors of this Fock space. More specifically,  `sector 1 of Fock space', that is, an individual Hilbert space, models `conceptual emergence', hence the combination of two concepts is represented by a superposition vector of 
the vectors representing the component concepts in this Hilbert space, allowing `quantum interference' between conceptual entities
to play a role in the process of emergence. `Sector 2 of Fock space', that is, a tensor product of two
versions of this Hilbert 
space, models a conceptual combination from the combining concepts by requiring 
the rules of logic 
for the logical connective used for the combining, i.e. conjunction or disjunction, 
to be satisfied in a probabilistic setting. This quantum-theoretic modeling suggested us to call `quantum conceptual thought' the process occurring in sector 1 of Fock space, `quantum logical thought' the process occurring in sector 2. \\
The relative importance of emergence 
or logic
in a specific cognitive process is measured by the `degree of participation' of sectors 1 and 2, as it will be clear in the next sections. 
The abundance of evidence of deviations from classical logical reasoning in concrete human decisions (paradoxes, fallacies, effects, contradictions), 
together with our results, led us to draw the conclusion that emergence constitutes the dominant dynamics of human reasoning, while logic is only a secondary form of dynamics. Hence, with respect to the aforementioned deviations from classicality, we 
put forward the view that what has been called a fallacy, an effect, a deviation, or a contradiction, is a consequence of the dominant dynamics and its nature is emergence, while what has been considered as a default to deviate from, namely classical logical reasoning, is a consequence of a secondary form
of dynamics, its nature being logic. This claim will be sustained and justified in the following sections, where we will apply our quantum-mechanical framework to model specific conceptual combinations and the respective experimental data. 
We also identify aspects of the history and nature of human though explaining the value with respect to the evolution of the human species of the presence of the two types of reasoning, logical reasoning with maximum value for situations about macroscopical physical entities and their dynamics and interactions, and emergent reasoning more proper for situations made of conceptual entities and their interactions, e.g. like they arise from both human minds during a conversation.     

\section{Modeling conceptual conjunction and disjunction in Fock space\label{conjdisj}} 

We present here our quantum modeling approach in Fock space for the conjunction and the disjunction of two concepts.
In the case of two combining entities, a Fock space $\mathcal F$ consists of two sectors: `sector 1' is a Hilbert space $\cal H$, while `sector 2' is a tensor product $\cal H \otimes \cal H$
of two isomorphic versions of $\cal H$.

Let us now consider the membership weights of items of concepts and their conjunctions/disjunctions measured by Hampton   \cite{h1988a,h1988b}. He identified systematic deviations from classical set (fuzzy set) conjunctions/disjunctions, an effect known as `overextension' or `underextension'.  

Let us 
firstly consider conjunctions. It can be shown that a large part of Hampton's data cannot be modeled in a classical probability space satisfying the axioms of Kolmogorov. For example, Hampton estimated the membership weight of  the item {\it Mint} with respect to the concepts {\it Food}, {\it Plant} and their conjunction {\it Food And Plant}, finding $\mu_{Mint}(Food)=0.87$, $\mu_{Mint}(Plant)=0.81$, $\mu_{Mint}(Food \ And \ Plant)=0.9$, respectively 
\cite{h1988a}. Thus, the item \emph{Mint} presents overextension with respect to the conjunction \emph{Food And Plant} of the concepts \emph{Food} and \emph{Plant}, and no classical probability representation exists for these data. 

Let us now come to disjunctions. Also in this case, a large part of Hampton's data cannot be modeled in a classical Kolmogorovian probability space. For example, Hampton estimated the membership weight of the item 
{\it Sunglasses} with respect to the concepts  
{\it Sportswear}, {\it Sports Equipment} and their disjunction 
{\it Sportswear Or Sports Equipment}, finding 
$\mu_{Sunglasses}(Sportswear)=0.4$, $\mu_{Sunglasses}(Sports \ Equipment)=0.2$, $\mu_{Sunglasses}(Sportswear \ Or \ Sports \ Equipment)=0.1$. Thus, the item 
\emph{Sunglasses} presents underextension with respect to the disjunction \emph{Sportswear Or Sports Equipment} of the concepts \emph{Sportswear} and \emph{Sports Equipment}, and no classical probability representation exists for these data 
\cite{h1988b}. 

It can be proved that a quantum probability model in Fock space exists for Hampton's data 
on conjunction and disjunction \cite{a2009a,ags2013}. Let us again start with the conjunction of two concepts. Let $x$ be an item and let $\mu(A)$, $\mu(B)$ and $\mu(A \ {\rm and} \ B)$  be the membership weights of $x$ with respect to the concepts $A$, $B$ and  `$A \ \textrm{and} \ B$' respectively. Let ${\cal F}={\cal H} \oplus ({\cal H} \otimes {\cal H})$ be the Fock space where we represent the conceptual entities. The states of the concepts $A$, $B$ and $`A \ \textrm{and} \ B'$ are represented by the unit vectors $|A\rangle, |B\rangle \in {\cal H}$ and $|A \ \textrm{and} \ B\rangle \in {\cal F}$, respectively, where
\begin{eqnarray}
|A \  \textrm{and} \ B\rangle=m e^{i\lambda}|A\rangle\otimes|B\rangle+ne^{i\nu}{1\over \sqrt{2}}(|A\rangle+|B\rangle)
\end{eqnarray}
The superposition vector ${1 \over \sqrt{2}}(|A\rangle+|B\rangle)$ describes `$A$ and $B$' as a new emergent concept, while the product vector $|A\rangle\otimes|B\rangle$ describes `$A$ and $B$' in terms of concepts $A$ and $B$. The positive numbers $m$ and $n$ are such that $m^{2}+n^{2}=1$. The decision measurement of a subject who estimates the membership of the item $x$ with respect to the concept `$A \  \textrm{and} \ B$' is represented by the orthogonal projection operator $M\oplus (M \otimes M)$ on ${\cal F}$, where $M$ is an orthogonal projection operator on ${\cal H}$. Hence, the membership weight of $x$ with respect to `$A \  \textrm{and} \ B$' is given by
\begin{eqnarray} \label{AND}
\mu(A \ \textrm{and} \ B)&=&\langle A \ \textrm{and} \ B|M \oplus (M \otimes M)|A \ \textrm{and} \ B \rangle \nonumber \\
&=&m^2\mu(A)\mu(B)+n^2 \left ( {\mu(A)+\mu(B) \over 2}+\Re\langle A|M|B\rangle \right )
\end{eqnarray}
where $\mu(A)=\langle A|M|A\rangle$ and $\mu(B)=\langle B|M|B\rangle$. The term $\Re\langle A|M|B\rangle$ is the `interference term' of quantum theory. A solution of Eq. (\ref{AND}) exists where this interference term given by
\begin{equation}
\Re\langle A|M|B\rangle=
\left\{
\begin{array}{ccc}
\sqrt{1-\mu(A)}\sqrt{1-\mu(B)}\cos\theta & & {\rm if} \  \mu(A)+\mu(B)>1 \\
\sqrt{\mu(A)}\sqrt{\mu(B)}\cos\theta  & & {\rm if} \ \mu(A)+\mu(B)\le 1
\end{array}
\right.
\end{equation}
($\theta$ is the `interference angle'). Coming to the example above, namely, the item {\it Mint} with respect to {\it Food}, {\it Plant} and {\it Food And Plant}, we have that Eq. (\ref{AND}) is satisfied with $m^2=0.3$, $n^2=0.7$ and $\theta=50.21^{\circ}$ (cfr., Ref. \cite{ags2013}, Sect. 3).

Keeping in mind the explanation we have given in Sect. \ref{nature}, we interpret this result on conceptual conjunction as follows. Whenever a subject is asked to estimate whether a given item $x$ belongs to the vague concepts $A$, $B$, $`A \ {\rm and} \ B'$, two mechanisms act simultaneously and in superposition in the subject's thought. A `quantum logical thought', which is a probabilistic version of the classical logical reasoning, where the subject considers two copies of item $x$ and estimates whether the first copy belongs to $A$ and the second copy of $x$ belongs to $B$,
and further the probabilistic version of the conjunction is applied to both estimates. But also a `quantum conceptual thought' acts, where the subject estimates whether the item $x$ belongs to the newly emergent concept `$A \ {\rm and} \ B$'.
The place whether these superposed processes can be suitably structured is Fock space. Sector 1 hosts the latter process, while sector 2 hosts the former, while the weights $m^2$ and $n^2$ measure the `degree of participation' of sectors 2 and 1, respectively, in the case of conjunction. 
In the case of {\it Mint}, subjects consider {\it Mint} to be more strongly a member of the concept {\it Food And Plant}, than they consider it to be a member of {\it Food} or of {\it Plant}. This is an effect due to a strong presence of quantum conceptual thought, the newly formed concept {\it Food And Plant} being found to be a better fitting category for {\it Mint} than the original concepts {\it Food} or {\it Plant}. 
And indeed, in the case of {\it Mint}, considering the values of $n^2$ and $m^2$, the combination process mainly occurs in sector 1 of Fock space, which means that emergence prevails over logic.

Consider instead a situation where logical aspects are prevalent over emergent aspects in our quantum-theoretic modeling. This situation is an example of a `borderline contradiction'. Suppose that a large sample of human subjects is asked to estimate the truth values of the sentences ``John is tall'', ``John is not tall'' and ``John is tall and not tall'', for a given subject John showed to the eyes of the subjects. And suppose that the fractions of positive answers are $0.01$,  $0.95$ and $ 0.15$, respectively \cite{ap2011}. This `borderline case' is clearly problematical from a classical logical perspective, and can be modeled in terms of overextension. Indeed, 
let us denote by $\mu(A)$, $\mu(A')$ and $\mu (A \ {\rm and} \ A')$ the probabilities that the sentences ``John is tall'', ``John is not tall'' and ``John is tall and not tall'' are true, and interpret them as membership weights of the item {\it John} with respect to the concepts {\it Tall}, {\it Not Tall} and {\it Tall And Not Tall}, respectively. Then Eq. ({\ref{AND}}) is solved for $m^{2}=0.77$, $n^{2}=0.23$ and $\theta=0^{\circ}$ \cite{s2014a}. The explanation of this behaviour is that the reasoning process of the subject mainly occurs in sector 2 of Fock space, hence 
logical reasoning is dominant, although emergent reasoning is also present, and it is its presence which evoked the name `contradiction' for this situation.

Let us now come to the disjunction of two concepts. Let $x$ be an item and let $\mu(A)$, $\mu(B)$ and $\mu(A \ {\rm or} \ B)$ be the membership weights of $x$ with respect to the concepts $A$, $B$ and `$A \ \textrm{or} \ B$', respectively. Let ${\cal F}={\cal H} \oplus ({\cal H} \otimes {\cal H})$ be the Fock space where we represent the conceptual entities. The concepts $A$, $B$ and `$A \ \textrm{or} \ B$' are represented by the unit vectors $|A\rangle, |B\rangle \in {\cal H}$ and $|A \ \textrm{or} \ B\rangle \in {\cal F}$, respectively, where
\begin{eqnarray}
|A \  \textrm{or}  \ B \rangle=m e^{i\lambda}|A\rangle\otimes|B\rangle+ne^{i\nu}{1\over \sqrt{2}}(|A\rangle+|B\rangle)
\end{eqnarray}
The superposition vector ${1 \over \sqrt{2}}(|A\rangle+|B\rangle)$ describes `$A$ or $B$' as a new emergent concept, while the product vector $|A\rangle\otimes|B\rangle$ describes `$A$ or $B$' in terms of concepts $A$ and $B$. The positive numbers $m$ and $n$ are such that $m^{2}+n^{2}=1$, and they estimate the `degree of participation' of sectors 2 and 1, respectively, in the disjunction case. The decision measurement of a subject who estimates the membership of the item $x$ with respect to the concept `$A \  \textrm{or} \ B$' is represented by the orthogonal projection operator $M \oplus ( M \otimes \mathbbmss{1}+\mathbbmss{1}\otimes M - M \otimes M)$ on ${\cal F}$, where $M$ has been introduced above. We observe that
\begin{equation}
 M \otimes \mathbbmss{1}+\mathbbmss{1}\otimes M - M \otimes M= \mathbbmss{1}-
 (\mathbbmss{1}-M)\otimes(\mathbbmss{1}-M)
\end{equation}
that is, in the transition from conjunction to disjunction we have applied de Morgan's laws of logic in sector 2 of Fock space. The membership weight of $x$ with respect to `$A \  \textrm{or} \ B$' is given by
\begin{eqnarray} \label{OR}
\mu(A \ \textrm{or} \ B)=\langle A \ \textrm{or} \ B | M \oplus (M \otimes \mathbbmss{1}+\mathbbmss{1}\otimes M - M \otimes M)
|A \ \textrm{or} \ B \rangle \nonumber \\
m^2 \left (\mu(A)+\mu(B)-\mu(A)\mu(B) \right  )+n^2 \left ( {\mu(A)+\mu(B) \over 2}+\Re\langle A|M|B\rangle \right )
\end{eqnarray}
where $\mu(A)=\langle A|M|A\rangle$ and $\mu(B)=\langle B|M|B\rangle$. The term $\Re\langle A|M|B\rangle$ is the interference term. A solution of Eq. (\ref{OR}) exists where this interference term given by
\begin{equation}
\Re\langle A|M|B\rangle=
\left\{
\begin{array}{ccc}
\sqrt{1-\mu(A)}\sqrt{1-\mu(B)}\cos\theta & & {\rm if} \  \mu(A)+\mu(B)>1 \\
\sqrt{\mu(A)}\sqrt{\mu(B)}\cos\theta  & & {\rm if} \ \mu(A)+\mu(B)\le 1
\end{array}
\right.
\end{equation}
Coming to the example above, namely, the item 
{\it Sunglasses} with respect to {\it Sportswear}, {\it Sports Equipment} and {\it Sportswear Or Sports Equipment}, we have that Eq. (\ref{OR}) is satisfied with $m^2=0.03$, $n^2=0.97$ and $\theta=155.00^{\circ}$.

Keeping in mind the explanation we have given in Sect. \ref{nature}, we interpret this result on conceptual disjunction as follows. Whenever a subject is asked to estimate whether a given item $x$ belongs to the vague concepts $A$, $B$, $`A \ {\rm or} \ B'$, two mechanisms act simultaneously and in superposition in the subject's thought. A `quantum logical thought', which is a probabilistic version of the classical logical reasoning, where the subject considers two copies of item $x$ and estimates whether the first copy belongs to $A$ or the second copy of $x$ belongs to $B$, and further the probabilistic version of the disjunction is applied to both estimates. And also a `quantum conceptual thought' acts, where the subject estimates whether the item $x$ belongs to the newly emergent concept `$A \ {\rm or} \ B$'. The place whether these superposed processes are structured is again Fock space. Sector 1 hosts the latter process, while sector 2 hosts the former, while the weights $m^2$ and $n^2$ measure the `degree of participation' of sectors 2 and 1, respectively, in the case of disjunction. 
In the case of {\it Sunglasses}, subjects consider {\it Sunglasses} to be less strongly a member of the concept {\it Sportswear Or Sports Equipment}, than they consider it to be a member of {\it Sportswear} or of {\it Sports Equipment}. This is an effect due to a strong presence of quantum conceptual thought, the newly formed concept {\it Sportswear Or Sports Equipment} being found to be a less well fitting category for {\it Sunglasses} than the original concepts {\it Sportswear} or {\it Sports Equipment}. 
And indeed, in the case of {\it Sunglasses}, considering the values of $n^2$ and $m^2$, the combination process mainly occurs in sector 1 of Fock space, which means that emergence aspects prevails over logical aspects in the reasoning process.

\section{Extending Fock space modeling to conceptual negation\label{CC-Neg}}
The first studies on the negation of natural concepts were also performed by Hampton \cite{h1997}. He tested membership weights on conceptual conjunctions  of the form {\it Tools Which Are Not Weapons} in experiments on human subjects,
finding overextension
and deviations from
Boolean behaviour in the negation.  More recently, we have performed a more general cognitive test inquiring into the membership weights of items with respect to conjunctions of the form {\it Fruits And Vegetables}, {\it Fruits And Not Vegetables}, {\it Not Fruits And Vegetables} and {\it Not Fruits And Not Vegetables} \cite{s2014b,asv2014b}. Our
   data confirmed significant deviations from classicality
and evidenced a very stable pattern of such deviations to the classicality conditions. The data could very faithfully be represented in two-sector Fock space, thus providing support to our quantum-theoretic modeling in Sect. \ref{conjdisj}. More, they allowed us to attain new fundamental results in concept research and to sustain and corroborate our explanatory hypothesis in Sect. \ref{nature}. Hence, it is worth to briefly review our recent results starting from the conditions for classicality of conceptual data sets, i.e. representability of empirical membership weights in a 
 Kolmogorovian probability space.
 
Let $\mu(A), \mu(B), \mu(A'), \mu(B')$, $\mu(A\ {\rm and}\ B)$, $\mu(A\ {\rm and}\ B')$, $\mu(A'\ {\rm and}\ B)$, and $\mu(A'\ {\rm and}\ B')$ be the membership weights of an item $x$ with respect to the concepts $A$, $B$, their negations `not $A$', `not $B$' and the conjunctions `$A$ and $B$', `$A$ and not $B$', `not $A$ and $B$' and `not $A$ and not $B$', respectively, and suppose that all these membership weights are  contained in the interval $[0,1]$ (which they will be in case they are experimentally determined as limits of relative frequencies of respective memberships). Then, they are `classical conjunction data' if and only if they satisfy the following conditions.
\begin{eqnarray} \label{condbis01}
&\mu(A)=\mu(A\ {\rm and}\ B)+\mu(A \ {\rm and}\ B') \\ \label{condbis02}
&\mu(B)=\mu(A\ {\rm and}\ B)+\mu(A' \ {\rm and}\ B) \\ \label{condbis03}
&\mu(A')=\mu(A'\ {\rm and}\ B')+\mu(A' \ {\rm and}\ B) \\ \label{condbis04}
&\mu(B')=\mu(A'\ {\rm and}\ B')+\mu(A \ {\rm and}\ B') \\ \label{condbis05}
&\mu(A\ {\rm and}\ B)+\mu(A\ {\rm and}\ B')+\mu(A'\ {\rm and}\ B)+\mu(A'\ {\rm and}\ B')=1
\end{eqnarray}
(see \cite{asv2014b} for the proof).

A large amount of data collected in \cite{asv2014b} violates very strongly and also very systematically Eqs. (\ref{condbis01})--(\ref{condbis05}), hence these data cannot be generally reproduced in a classical Kolmogorovian probability framework. It can instead be shown that almost all these data can be represented by using our  quantum-theoretic modeling in two-sector Fock space, as in Sect. \ref{conjdisj}. For the sake of simplicity, let us work out separate representations for the two sectors.

Let us start from sector 1 of Fock space, which models genuine emergence. We represent the concepts $A$, $B$ and their negations `not $A$', `not $B$' by the mutually orthogonal unit vectors $|A\rangle$, $|B\rangle$ and $|A'\rangle$, $|B'\rangle$, respectively, in the individual Hilbert space ${\cal H}$. The corresponding membership weights for a given item $x$ are then given by the quantum probabilistic Born rule
\begin{eqnarray}
\mu(A)=\langle A|M|A\rangle & \quad & \mu(B)=\langle B|M|B\rangle \\
\mu(A')=\langle A'|M|A'\rangle& \quad & \mu(B')=\langle B'|M|B'\rangle
\end{eqnarray}
in sector 1. The conjunctions `$A$ and $B$', `$A$ and not $B$', `not $A$ and $B$', and `not $A$ and not $B$' are represented by the superposition vectors $\frac{1}{\sqrt{2}}(|A\rangle+|B\rangle)$, $\frac{1}{\sqrt{2}}(|A\rangle+|B'\rangle)$, $\frac{1}{\sqrt{2}}(|A'\rangle+|B\rangle)$ and $\frac{1}{\sqrt{2}}(|A'\rangle+|B'\rangle)$, respectively, in ${\cal H}$, i.e. sector 1 of Fock space, which expresses the fact `$A$ and $B$', `$A$ and not $B$', `not $A$ and $B$', and `not $A$ and not $B$' are considered as newly emergent concepts in sector 1.

Let us come to sector 2 of Fock space, which models logical reasoning.  
Here we introduce a new element,
expressing an insight which we had not 
yet in our earlier application of Fock space \cite{a2009a,ags2013,s2014b,s2014a}, and which we explain in detail in \cite{asv2014b}. In short it comes to `taking into account that possibly $A$ and $B$ are meaning-connected and hence their probability weights mutually 
dependent'. If this is the case, we cannot represent, e.g., the conjunction `$A$ and $B$' by the tensor product vector $|A\rangle \otimes |B\rangle$ of ${\cal H} \otimes {\cal H}$, as we have supposed in Sect. \ref{conjdisj}. This would indeed entail that the membership weight for the conjunction is $\mu(A \ {\rm and}  \ B)=\mu(A)\mu(B)$ in sector 2, that is, probabilistic independence between the membership estimations of $A$ and $B$. We instead, following this new insight, represent the conjunction `$A$ and $B$' by an arbitrary vector $|C\rangle\in {\cal H} \otimes {\cal H}$, in sector 2, which in general will be entangled if $A$ and $B$ are meaning dependent. If we represent the decision measurements of a subject estimating the membership of the item $x$ with respect to the concepts $A$ and $B$ by the orthogonal projection operators $M\otimes \mathbbm{1}$ and $\mathbbm{1}\otimes M$, respectively, we have
\begin{equation}
\mu(A)=\langle C|M\otimes \mathbbm{1}|C\rangle \quad
\mu(B)=\langle C| \mathbbm{1}\otimes M|C\rangle
\end{equation}
in sector 2. We have now to formalize the fact that this sector 2 has to express logical relationships between the concepts. More explicitly, the decision measurements of a subject estimating the membership of the item $x$ with respect to the negations `not $A$' and `not $B$' should be represented by the orthogonal projection operators $(\mathbbmss{1}-M)\otimes \mathbbm{1}$ and $\mathbbm{1}\otimes (\mathbbmss{1}-M)$, respectively, in sector 2, in such a way that
\begin{equation}
\mu(A')=1-\mu(A)=\langle C|(\mathbbm{1}-M)\otimes \mathbbm{1}|C\rangle \quad
\mu(B')=1-\mu(B)=\langle C| \mathbbm{1}\otimes (\mathbbm{1}-M|C\rangle)
\end{equation}
in this sector.

Interestingly enough, there is a striking connection between logic and classical probability when conjunction and negation of concepts are at stake. Namely, the logical probabilistic structure of sector 2 of Fock space sets the limits of classical probabilistic models, and  vice versa. In other words, if the experimentally collected membership weights $\mu(A)$, $\mu(B)$, $\mu(A')$, $\mu(B')$, $\mu(A\ {\rm and}\ B)$, $\mu(A \ {\rm and}\ B')$, $\mu(A'\ {\rm and}\ B)$ and $\mu(A'\ {\rm and}\ B')$ can be represented in sector 2 of Fock space for a given choice of the  state vector $|C\rangle$ and the decision measurement projection operator $M$, then the membership weights satisfy (\ref{condbis01})--(\ref{condbis05}), hence they are classical data.  Vice versa, if $\mu(A)$, $\mu(B)$, $\mu(A')$, $\mu(B')$, $\mu(A\ {\rm and}\ B)$, $\mu(A \ {\rm and}\ B')$, $\mu(A'\ {\rm and}\ B)$ and $\mu(A'\ {\rm and}\ B')$ satisfy (\ref{condbis01})--(\ref{condbis05}), hence they are classical data, then an entangled state vector $|C\rangle$ and a decision measurement projection operator $M$ can always be found such that $\mu(A)$, $\mu(B)$, $\mu(A')$, $\mu(B')$, $\mu(A\ {\rm and}\ B)$, $\mu(A \ {\rm and}\ B')$, $\mu(A'\ {\rm and}\ B)$ and $\mu(A'\ {\rm and}\ B')$ can be represented in sector 2 of Fock space (see \cite{asv2014b} for the proof).

Let us finally come to the general representation in two-sector Fock space. We can now introduce the general form of the vector representing the state of the conjunction of the concepts $A, B$ and their respective negations.
\begin{eqnarray}
| \Psi_{AB} \rangle&=& m_{AB}e^{i \lambda_{AB}} |C\rangle + \frac{n_{AB}e^{i \nu_{AB}}}{\sqrt{2}} (|A\rangle+|B\rangle) \\
| \Psi_{AB'} \rangle&=& m_{AB'}e^{i \lambda_{AB'}} |C\rangle + \frac{n_{AB'}e^{i \nu_{AB'}}}{\sqrt{2}} (|A\rangle+|B'\rangle) \\
| \Psi_{A'B} \rangle&=& m_{A'B}e^{i \lambda_{A'B}} |C\rangle + \frac{n_{A'B}e^{i \nu_{A'B}}}{\sqrt{2}} (|A'\rangle+|B\rangle) \\
| \Psi_{A'B'} \rangle&=& m_{A'B'}e^{i \lambda_{A'B'}} |C\rangle + \frac{n_{A'B'}e^{i \nu_{A'B'}}}{\sqrt{2}} (|A'\rangle+|B'\rangle)
\end{eqnarray}
where $m^2_{XY}+n^2_{XY}=1$, $X=A,A',Y=B,B'$. 
The corresponding membership weights are 
\begin{eqnarray}
\mu(A\ {\rm and}\ B)&=&m_{AB}^2 \alpha_{AB}+n_{AB}^2({1 \over 2}(\mu(A)+\mu(B))+\beta_{AB}\cos\phi_{AB})
\label{FockSpaceSolutionAB} \\
\mu(A \ {\rm and} \ B')&=&m_{AB'}^2 \alpha_{AB'}+n_{AB'}^2({1 \over 2}(\mu(A)+\mu(B'))+\beta_{AB'}\cos\phi_{AB'}) \label{FockSpaceSolutionAB'} \\
\mu(A' \ {\rm and}\ B)&=&m_{A'B}^2 \alpha_{A'B}+n_{A'B}^2({1 \over 2}(\mu(A')+\mu(B))+\beta_{A'B}\cos\phi_{A'B})
\label{FockSpaceSolutionA'B} \\
\mu(A'\ {\rm and}\ B')&=&m_{A'B'}^2\alpha_{A'B'}+n_{A'B'}^2({1 \over 2}(\mu(A')+\mu(B'))+\beta_{A'B'}\cos\phi_{A'B'}) \label{FockSpaceSolutionA'B'}
\end{eqnarray}
where $0\le \alpha_{XY},\beta_{XY}\le 1$,  $X=A,A',Y=B,B'$.

Let us consider a relevant example, {\it Goldfish}, with respect to ({\it Pets}, {\it Farmyard Animals}) (big overextension in all experiments, but also double overextension with respect to {\it Not Pets And Farmyard Animals}). {\it Goldfish} scored $\mu(A)=0.93$ with respect to {\it Pets}, $\mu(B)=0.17$ with respect to {\it Farmyard Animals}, $\mu(A')=0.12$ with respect to {\it Not Pets}, $\mu(B')=0.81$ with respect to {\it Not Farmyard Animals}, $\mu(A \ {\rm and} \ B)=0.43$ with respect to {\it Pets And Farmyard Animals}, $\mu(A \ {\rm and} \ B')=0.91$ with respect to {\it Pets And Not Farmyard Animals}, $\mu(A' \ {\rm and} \ B)=0.18$ with respect to {\it Not Pets And Farmyard Animals}, and $\mu(A' \ {\rm and} \ B')=0.43$ with respect to {\it Not Pets And Not Farmyard Animals}. A complete modeling in the Fock space satisfying Eqs. (\ref{FockSpaceSolutionAB}), (\ref{FockSpaceSolutionAB'}), (\ref{FockSpaceSolutionA'B}) and (\ref{FockSpaceSolutionA'B'}) is characterized by coefficients:

(i) interference angles $\phi_{AB}=78.9^{\circ}$, $\phi_{AB'}=43.15^{\circ}$, $\phi_{A'B}=54.74^{\circ}$ and $\phi_{A'B'}=77.94^{\circ}$;

(ii) coefficents $\alpha_{AB}=0.35$, $\alpha_{AB'}=0.9$, $\alpha_{A'B}=0.22$ and $\alpha_{A'B'}=0.17$;

(iii) coefficients 
$\beta_{AB}=-0.24$, 
$\beta_{AB'}=0.10$, $\beta_{A'B}=0.12$ and $\beta_{A'B'}=0.30$;

(iv)  convex weights 
$m_{AB}=0.45$,
$n_{AB}=0.89$, 
$m_{AB'}=0.45$, 
$n_{AB'}=0.9$, 
$m_{A'B}=0.48$,  $n_{A'B}=0.88$, $m_{A'B'}=0.45$, and $n_{A'B'}=0.89$. 

\section{Origin and foundations of human reasoning\label{aristotle}}
In the previous sections we have provided concrete evidence that human reasoning is actually guided by two drivers, namely, emergence -- which stands for the continuous creation of new conceptual structures carrying also new meaning whenever the human mind combines pieces of existing conceptual structure containing given meaning -- and logic  -- which consists of applying the probabilistic rules of logic for conjunctions, disjunctions and negations appearing in these conceptual combinations. And that these aspects of human reasoning can be formalized by using the mathematical formalism of quantum theory in Fock space. We have also made clear that emergent reasoning generally prevails over logical reasoning in concrete human decisions. 
 In this respect, the systematic effects, fallacies and paradoxes discovered in experimental psychology should not be considered as biases of human reasoning but, 
  rather, as fundamental expressions of conceptual emergence at the deepest level. 

There is 
further empirical evidence revealing that what really guides human subjects in a concrete decision is not logical but emergent reasoning. Consider, for example, the item {\it Olive} and its membership weights with respect to the concepts {\it Fruits}, {\it Vegetables} and their disjunction {\it Fruits Or Vegetables}, measured by Hampton \cite{h1988b}, and its membership weights with respect to the concepts {\it Fruits}, {\it Vegetables} and their conjunction {\it Fruits And Vegetables}, measured by ourselves \cite{s2014b,asv2014b}. 
{\it Olive} scored $\mu(A)=0.5$ with respect to {\it Fruits}, $\mu(B)=0.1$ with respect to {\it Vegetables} and $\mu(A \ {\rm 
or} \ B)=0.8$ with respect to {\it Fruits Or Vegetables}, that is, {\it Olive} was double overextended with respect to the disjunction. In addition, {\it Olive} scored $\mu(A)=0.56$ with respect to {\it Fruits}, $\mu(B)=0.63$ with respect to {\it Vegetables} and $\mu(A \ {\rm and} \ B)=0.65$ with respect to {\it Fruits 
And Vegetables}, that is, {\it Olive} was double overextended also with respect to the 
conjunction. This means that 
situations exist where  
people do not really take into account whether the connective `or', or the connective `and', is considered, but they actually estimate whether the item $x$ is a member of the new emergent concept, be it `$A$ or $B$', or `$A$ and $B$'. Indeed, the membership weight oscillates around the average $\frac{1}{2}(\mu(A)+\mu(B))$, due to the interference term. But, sector 1 emergence almost completely drives the cognitive dynamics.

Now, since the origins of logic can be traced back to the ancient Greeks, while emergent phenomena are a relatively recent discovery of modern science, one is naturally led to inquire into the deep motivations why logical reasoning has been historically formalized much before emergent reasoning. Why did logic appear first? We think that there are two main reaons for this, as follows.

(i) Emergent reasoning, as formalized in sector 1 of Fock space, needs sophistiacted technical tools, as 
 algebraic structures, Hilbert spaces, etc., while logical reasoning can basically be formalized by combining more intuitive structures. It is then not implausible that the first set-theoretical models by ancient Greeks could better capture the latter than the former reasoning.

(ii) Ancient Greeks formalized logical reasoning with the objective of fully understanding mathematical reasoning. Since then, mathematics develops a strong method of reasoning (the one formalized in sector 2 of Fock space), and limits its application range to the domains where such a mathematical reasoning is applicable.

Point (ii) is important, 
 because it helps understanding why the effects discovered by Tversky, Khaneman and other cognitive psychologists were first classified as biases  or even fallacies of human reasoning. If mathematical reasoning is the most rigorous guide of classical logical reasoning, then each deviation from mathematical reasoning, e.g., wrong application of probabilities in concrete human judgements, should be considered as a real fallacy of human reasoning. 

This leads us to another interesting question. Which are the domains of our world where logical reasoning is more valuable or perhaps even the only valuable type of reasoning, such as in mathematics? And, which are the domains of our world where emergent reasoning is the more valuable type of reasoning to be used? We believe that, whenever reasoning has as subject 
`concrete material objects being present in real space-like configuration', logical reasoning is the one which is more appropriate. People who give dominance to emergent reasoning in these situations 
defined mainly by presences of material objects in space and  time will commit fallacies of reasoning, and possibly their reasoning will lead to expectations about things to happen in this material object world that do not happen following these expectations. The reason is that this material world of objects in space-time can be modelled by considering sets, subsets, movements from one set into another, hence a dynamics of classical physics nature. Of course, for real situations such a description might be very complex, but for sector 2 Fock space reasoning to be valid, it is sufficient that in principle 
such a classical mechanics model is possible. To give 
an easier picture of what we mean, manuals that exist as guides for the functioning of technical mechanical or electric devices are good examples of descriptions which allow such sector 2 of Fock space logical reasoning. So, the question `whether a specific type of happenings can be captured completely in a manual' is a good criterion for 
sector 2 Fock space reasoning validity. And, what are the domains of our world where 
sector 1 Fock space reasoning is the more valid one? We believe that, whenever the subject of our reasoning consists of complex enough conceptual situations, the human mind is in need of the emergence of new meaning and new conceptual structures at every instant. It is what we call `creativity'. And again, in case in these domains where emergent conceptual reasoning is the 
most valid one people prefer to give dominance to sector 2 Fock space logical reasoning, this will not necessarily be experienced as fallacies but, rather, as `nerdy type of not understanding'. To state it metaphorically, `A poem is not a manual'.

Again to mathematics, is this not a discipline of reasoning focused completely on conceptual entities and not at all on material objects? It is indeed. But, in some way, mathematicians have delimited their subject material in function of 
sector 2 
logical reasoning being applicable to it, and even a strictly deterministic version of it. We believe that, first the liar paradox, and 
then G\"odel's paradox, and all its equivalents, e.g., the halting problem,
are situations where the systematics of applying classical logic failed also in the mathematical realm. It is most probably not a coincidence that quite some time ago
we produced in Brussels a quantum model description for the cognitive dynamics of the liar paradox situation \cite{aertsbroekaertsmets1999}.

To conclude, we have unveiled in this paper crucial new aspects of the foundations of human reasoning with our explanatory 
hypothesis on the existence of two layers in human thought guiding cognitive dynamics. However, we believe that still other aspects of the nature of human reasoning remain to be identified,
and we plan to perform new experiments to deepen this fascinating argument in future research.

\end{document}